\documentclass[runningheads]{llncs}
\usepackage[T1]{fontenc}
\usepackage{hyperref}
\usepackage{amsmath,amsfonts,amssymb}
\usepackage{graphicx,cite,enumerate}
\usepackage[ruled,longend]{algorithm2e}
\usepackage{textgreek}
\usepackage{amssymb}
\usepackage{tabularray}
\begin{document}
\title{Question Answering with Texts and Tables through Deep Reinforcement Learning}
\titlerunning{QA with Texts and Tables through Deep Reinforcement Learning}
%
\author{Marcos M. José\inst{1,\dagger}\orcidID{0000-0003-4663-4386} \and
Flávio N. Cação\inst{1,2}\orcidID{0000-0003-4771-6009} \and
Maria F. Ribeiro \and
Rafael M. Cheang\inst{1} \orcidID{0000-0003-2434-1304} \and
Paulo Pirozelli\inst{1} \orcidID{0000-0002-4714-287X} \and
Fabio G. Cozman\inst{1}\orcidID{0000-0003-4077-4935}
}

\authorrunning{M. José et al.}

\institute{Escola Politécnica, Universidade de São Paulo, São Paulo, Brazil \and Novelo Data, São Paulo, Brazil \\
$\dagger$ Correspondence to: \email{marcos.jose@alumni.usp.br}
}

%
\maketitle              
\begin{abstract}
This paper proposes a novel architecture to generate multi-hop answers to open domain questions that require information from texts and tables, using the Open Table-and-Text Question Answering dataset for validation and training.
One of the most common ways to generate answers in this setting is to retrieve information sequentially, where a selected piece of data helps searching for the next piece. 
As different models can have distinct behaviors when called in this sequential information search, a challenge is how to select models at each step. 
Our architecture employs reinforcement learning  to choose between different state-of-the-art tools sequentially until, in the end, a  desired answer is generated. 
This system achieved an F1-score of 19.03, comparable to iterative systems in the literature.

\end{abstract}

\section{Introduction}
Question Answering (QA) is concerned with developing programs that can respond to questions posed in natural language. Those systems are widely used to assist users with their queries without the need for direct human intervention. QA datasets typically contain question-answer pairs written in natural language; e.g., SQuAD \cite{SQUAD20216}. While textual data is the most common source of information, other formats, such as graphs \cite{Berant2013SemanticPO} and tabular data \cite{chen2020_hybridqa}, may contribute in different ways to the task. Tabular data is particularly valuable for structured information, which can be challenging to retrieve in pure textual databases.  A mix of textual and tabular data, thus, offers significant potential for improvement in QA systems.

This work addresses  QA problems that demand retrieving and aggregating information from both tabular and textual data. Our testbed is the Open Table-and-Text Question Answering (OTT-QA) dataset \cite{ott_qa}, an open-domain multi-hop QA dataset. In this dataset, questions are provided without supporting texts, and to answer them, one is typically required to navigate through multiple passages, textual and tabular. Performance on this dataset remains limited compared to purely textual datasets like SQuAD, where trained models already outperform humans. 

Currently, the most adopted strategies to QA combine retrieval and reader modules: a retriever module fetches relevant passages, whereas a reader module turns these passages into a meaningful answer \cite{Chen17,guu2020realm,RAG}. In multi-hop datasets like OTT-QA, this retrieval process needs to be done sequentially, moving from a piece of information to another, in order to get the required data (textual or tabular). Even though recent non-sequential systems have attained reasonable performance, a sequential decision-making formulation seems to be essential to coordinate retriever and reader schemes. 

In this paper we wish to explore 
the boundaries of performance obtained with 
sequential decisions applied to standard modules, where we resort to Reinforcement 
Learning (RL) to learn strategies.
 RL is suitable for this task: while supervised learning depends on labeled, static data for the action sequences, in RL the sequence of decisions is built through evolving information \cite{tese_flavio}.
We thus propose a novel Deep Reinforcement Learning (DRL) approach for this task, which iteratively decides which actions to take (retrieve text, retrieve table, generate answer), using standard modules to conduct these actions.\footnote{Data and code available at: \url{https://github.com/MMenonJ/DRL_QA_TT}}

The paper is organized as follows. Section \ref{section:Background} offers some necessary background. 
Section \ref{chapter:proposed_architecture} presents our proposed RL architecture.
Section \ref{chapter:experiments}  describes our experimental setup, while
section \ref{chapter:results} presents the results and their associated discussion. 
Finally, section \ref{chapter:conclusion} presents our concluding thoughts. 

\section{Background}\label{section:Background}

\subsection{Open Table-and-Text Question Answering}

%
Numerous  QA datasets and architectures have been designed to address questions that require understanding textual information, such as SQuAD, or tabular data, as MIMICSQL \cite{MIMICSQL}. However, there are not many resources that handle both texts and tables. 
To bridge this gap, HybridQA has been introduced as a closed-domain multi-hop multimodal question-answering dataset \cite{chen2020_hybridqa}. In HybridQA, each question carries a table and multiple associated passages, and to answer a question, it is necessary to combine several of these sources together.

\begin{figure}[t]
    \centering
    \includegraphics[width=1\textwidth]{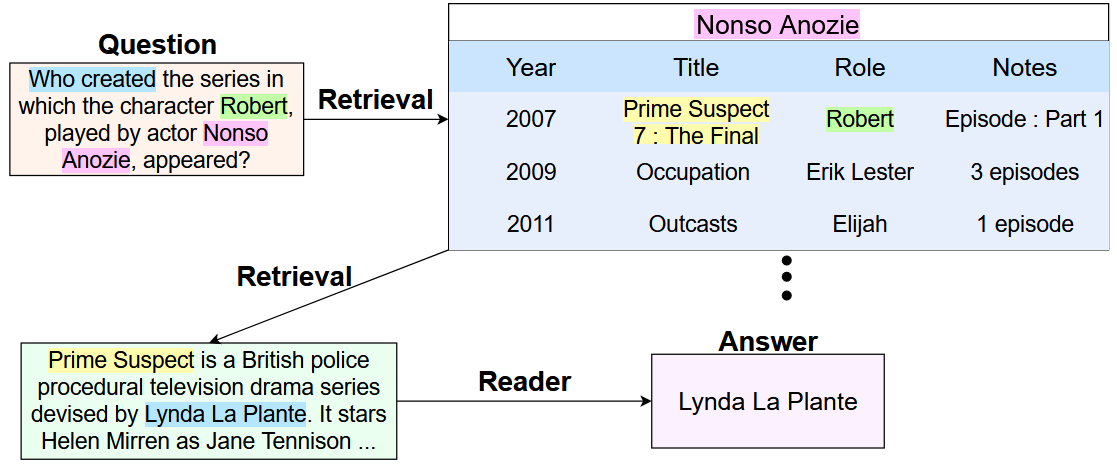}
    \caption[QA pair example taken from  OTT-QA \cite{ott_qa}]{QA pair from  OTT-QA \cite{ott_qa}. The question asks who is the author of the series in which Nonso Aznozie played the character Robert. To accurately answer this question, one must locate the relevant table labeled ``Nonso Anozie'' (highlighted in pink) and identify the cell with ``Prime Suspect'' (highlighted in yellow), by referencing the role ``Robert'' (highlighted in green). It is then possible to retrieve the associated text, which reveals that the creator of the movie was ``Lynda La Plante'' (highlighted in blue). }
    \label{fig:ott_qa}
\end{figure}

OTT-QA  \cite{ott_qa} is an English dataset with 45841 QA pairs built over HybridQA. 
Its core aim is to serve as an open-domain dataset that combines   text and tabular information while simultaneously increasing the complexity of the task. 
The dataset contains tables and texts with answers for every question in the dataset.
An example of a QA pair from OTT-QA is shown in Figure \ref{fig:ott_qa}.

To assess the answers, we employ two well-known evaluation metrics (popularized by the SQuAD dataset \cite{SQUAD20216} and used by OTT-QA): 
\textbf{Exact Match},  the percentage of   predicted answers that are exactly equal to the golden answers, and 
\textbf{Macro Average F1-score},  the overlap between   predicted and  golden answers. In this context, precision is  the ratio of  shared words between the golden and prediction answers over the total number of prediction words; and recall is the ratio of shared words over the number of words in the golden answer. The F1-score is the harmonic mean between the precision and the recall. 
 

\subsection{Open-Domain and Question Answering}

In open-domain QA, the questions might be about any topic and a system must find answers based on a given question and usually within a corpus of information. QA systems typically consist of two main components: a retriever and a reader. The retriever's task is to search the corpus for relevant texts and passages, while the reader processes this retrieved information to generate a suitable answer. Some approaches leverage neural transformers for both retriever and reader tasks \cite{RAG, guu2020realm}, whereas others use keyword-based models specifically for the retriever \cite{Chen17,bracis2021}.

\textbf{Retriever:} A widely adopted keyword-based retriever model in QA is BM25 \cite{Robertson2009}. BM25 relies on TF-IDF (Term Frequency–Inverse Document Frequency) by incorporating document length normalization; hence, shorter documents matching the query are prioritized over longer ones with similar keyword matches.
In contrast, Dense Passage Retrieval (DPR) \cite{karpukhin-etal-2020-dense} overcomes the limitations of keyword matching by ranking passages based on semantic similarity. DPR utilizes two BERT \cite{BERT} models trained jointly to produce dense vector representations of questions and passages --- similarity between these vectors is estimated by the dot product, enabling more nuanced retrieval based on meaning.



\textbf{Reader:} The best current readers typically are based on transformer neural networks, which take concatenated inputs of questions and retrieved passages to generate answers. For example, BERT-based models, as seen in Guu et al. \cite{guu2020realm}, focus on substring prediction, while encoder-decoder architectures like BART \cite{bart}, as in Lewis et al. \cite{RAG}, excel in text-to-text tasks.
Another notable reader  is the Fusion-in-Decoder (FiD) \cite{fusion_decoder}. Unlike simple concatenation of question and passages followed by direct use in an encoder-decoder network, FiD independently processes each passage in the encoder and combines resulting vectors as input to the decoder. This method efficiently handles a larger number of documents.
Similarly, Fusion-in-Encoder (FiE) \cite{fie} integrates information from multiple passages within the encoder rather than the decoder. FiE creates a unified representation and incorporates cross-sample attention over all tokens across different samples. The authors propose an alternative approach for calculating answer span probability to effectively consider information from all passages.

\subsection{Architectures tested on OTT-QA}\label{architectures_ott}
The authors of OTT-QA \cite{ott_qa} presented several solutions, each addressing different retrieval and reading strategies. The first solution, BM25-HYBRIDER, establishes a baseline for OTT-QA. It retrieves 1 to 4 texts or tables from the corpus using the BM25 algorithm. These retrieved passages are then input to a  reader, which selects the answer with the highest confidence interval.

The second solution, ``Iterative-Retrieval + Cross-Block Reader,'' employs two iterative retriever variants. \textbf{Sparse} uses BM25 to retrieve 10 text passages and 10 table segments. For each text passage, it retrieves 5 table segments, and vice versa. \textbf{Dense} utilizes a DPR-like model for retrieval. It starts with 8 text or table passages and iteratively expands the search. Both variants feed the retrieved passages into the Cross-Block Reader, a fine-tuned ETC transformer model \cite{ETC}, to generate the answer.

The last solution introduced by the authors, the ``Fusion Retriever + Cross-Block Reader'', was shown to outperform the previous methods. This approach combines tables and text into blocks using BM25 and enhances queries with GPT-2 \cite{GPT2}. The Fusion Retriever is a dual-encoder setup, based on DPR, that retrieves $k$ blocks based on their similarity to the question while maintaining the same Cross-Block Reader to generate the answer.

Other techniques have also been tested in OTT-QA, like the work of \citeonline{li2021dual}, which proposes a Retriever-Reader system and a Joint Reranking model known as DuRePa. The Retriever component utilizes BM25 to select 100 textual and 100 table passages relevant to the question. These passages are then fed into the Joint Reranking model, which employs a BERT-based architecture to assign scores to each passage, ultimately selecting the top 50 passages. The Reader component, based on a Fusion-in-Decoder approach with T5 \cite{2020t5}, determines whether the answer relies on a table, in which case it generates a SQL query, or if it is text-based, the model generates the final response.

The ChAincentric Reasoning and Pre-training framework (CARP) \cite{CARP}   builds text and passage blocks by utilizing early fusion techniques. This process involves employing a BERT model to establish entity links known as BLINK \cite{10.1145/3549737.3549771}, while a DPR, trained from RoBERTa \cite{roberta}, serves as the retriever for searching these blocks. Subsequently, CARP employs a Hybrid chain extractor, which is based on BART, to identify and extract links of knowledge. CARP's reader, based on  Longformer \cite{Longformer}, then leverages the assembled text blocks and knowledge chains to generate the answer.

OpenQA Table-Text Retriever (OTTER) \cite{otter} is another QA system tested on the OTT-QA dataset that  utilizes an early fusion method equal to CARP, and the Cross-Block Reader from OTT-QA's original paper. However, it makes significant improvements in the retriever. First, it enhances mixed-modality representation learning through modality-enhanced representation (MER) and mixed-modality hard negative sampling (MMHN). MER enriches the semantics by incorporating fine-grained representations of both tabular and textual data. MMHN, on the other hand, generates challenging hard negatives by substituting partial information within tables or texts to encourage better discrimination of relevant evidence. To overcome the data sparsity problem, the authors implement retrieval-centric mixed-modality synthetic pre-training. This involves constructing a large-scale synthesized corpus through mining relevant table-text pairs and generating pseudo questions using a BART model.

The CORE (Chain Of REasoning) QA system employs a retriever-reader architecture but includes two intermediary components, the Linker and the Chainer \cite{core}. The retriever is built on DPR and is responsible for retrieving 100 tables from the corpus. Subsequently, a Linker system based on BERT models retrieves passages from the corpus based on the rows obtained from the retrieved tables. As the linker may provide an excessive amount of information for the reader to process, the Chainer's role is to select the top 50 chains (table row and text passage), which consist of a table row and a corresponding text passage.
The Chainer accomplishes this task by employing a T0 \cite{t0} neural network in a zero-shot manner. It assesses the probabilities of the model generating a question based on specific text passages and table rows (the higher the probability of that question being generated, the higher the rank of the chain). Finally, to answer the question, the system uses a FiD model, which relies on the information curated by the Chainer.

The Chain-of-Skills (COS) architecture \cite{cos} currently delivers the highest performance on OTT-QA.
The authors introduced a (non-sequential) modularization approach that facilitates  multi-task training across various open-domain QA datasets. The tasks tackled include single retrieval, expanded query retrieval, entity span proposal, entity linking, and reranking.
In the inference phase, the system first  identifies 100 tables through single retrieval. These tables are subsequently divided into rows, and the reranking process selects the top 200 rows. For each row, the system collects 10 passages from the expanded query retrieval and one passage from linked entities. The system then leverages the same Chainer used in CORE to select the top 100 passages, which are subsequently input into a fine-tuned FiE reader tailored for OTT-QA.

Another significant contribution is the ``Multi-modal Retrieval of Tables and Texts Using Tri-encoder Models'' \cite{tri_encoder}, which extends DPR's capabilities by incorporating three small-BERT networks for encoding questions, texts, and tables individually. This architecture enhances neural retrieval efficiency for both tables and texts in QA systems.

\subsection{Reinforcement Learning in Question Answering}

One of the pioneering approaches to integrate Deep Reinforcement Learning (DRL) into QA is the Reinforced Ranker-Reader (\(R^3\)) architecture \cite{r3} introduced in 2017. This framework focuses on joint training of the retriever (referred to as the ranker) and the reader, using the REINFORCE algorithm for the ranker and a supervised approach for the reader. A key advantage of this approach is the unsupervised nature of retriever training, which gives rewards based solely on the similarity of the reader's output to the golden answer.

Similarly, MSCQA \cite{MSCQA} employs DRL to train an action selector that decides among three components: invoking the reader, selecting additional passages via the retriever, and removing potentially incorrect answers from memory. In \cite{tese_flavio}, a DRL agent is designed with actions for retriever, reader, and cleaner functions, in order to tackle a complex multi-hop QA dataset, where the retriever iteratively uses previously retrieved passages to find new relevant information.

An interesting proposal that uses Reinforcement Learning for automatic QA is NLPGYM by \citeonline{NLPGYM}. It is a Python library made for testing RL Algorithms in three different NLP tasks: multiple choice QA, label sequence generation, and the sequence tagging. The QA task is based on the QASC dataset \cite{QASC}, the state comprises the embedding of a question, two facts, and an alternative. The agent is equipped with two actions: answering using the provided alternative or requesting a new one. 

Reinforcement Learning with Human Feedback (RLHF) is an approach that has gained traction in the development of natural language models. 
RLHF involves training models through a process that combines Reinforcement Learning, where models learn from their actions and consequences, with Human Feedback, which involves human reviewers providing feedback and rankings on model-generated content. 
One example is ChatGPT (or GPT3.5), a conversational agent built upon the foundation of GPT3. While the specifics of ChatGPT's training process are not fully disclosed, available information indicates that the RLHF training was executed through the PPO \cite{PPO} algorithm to enhance ChatGPT's alignment with users and minimizing issues related to misinformation, toxicity, and harmful sentiments.

\section{Proposed Architecture}\label{chapter:proposed_architecture}

The multi-hop and multi-model nature of OTT-QA presents a significant challenge due to the sequential retrieval process it requires. To address these challenges within an explicitly sequential formulation, we propose a novel architecture. Our approach involves training a Deep Reinforcement Learning (DRL) agent that selects from a set of already trained modules in the existing literature.

\begin{figure}[t]
    \centering
    \includegraphics[width=0.7\textwidth]{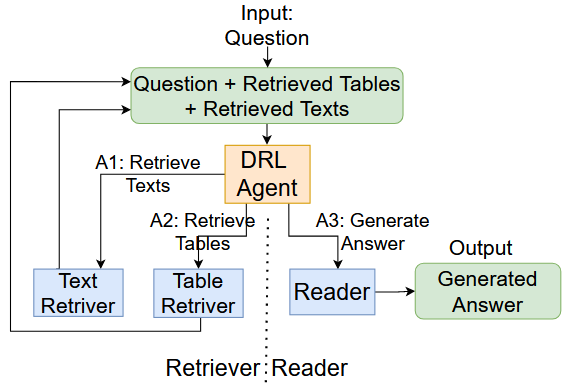}
    \caption[Proposed architecture.]{Proposed architecture. At each time step, the agent selects one of three actions: Retrieve Texts, Retrieve Tables, or Generate the Answer, based on the question and the information gathered so far.}
    \label{fig:Architecture}
\end{figure}

As done for MSCQA \cite{MSCQA}, we employ a DRL agent as an action selector to determine which pre-trained component should be activated. As depicted in Figure \ref{fig:Architecture}, we have the following actions: Retrieve Texts, Retrieve Tables, and Generate the Answer.

Given that OTT-QA does not specify the correct path for information retrieval, including the sequence of passages to retrieve, we have adopted a Reinforcement Learning approach. We incorporate a delayed reward mechanism, which compares the predicted answers to the gold standard answers at the end of each episode.

One notable advantage of our proposed architecture is its flexibility. All components, including the reader and retrievers, can be easily replaced with newer and superior models as they become available for public use. Furthermore, the system can be enhanced by incorporating additional components, such as a graph retrieval model, or employing multiple readers optimized for specific scenarios. The RL agent learns when to use each component in different situations through experiential training.


\subsection{Initial Concepts}
Our approach draws inspiration from the Iterative-Retrieval + Cross-Block Reader architecture \citeonline{ott_qa} described  in Section \ref{architectures_ott}, being also an iterative system. For a given question, the initial retrieval step involves searching for 10 passages, which can be either textual or tabular in nature. These passages are used to create 10 different blocks. 
Should the agent opt to perform an additional retrieval step, it concatenates the items within each block to retrieve four new passages, be they textual or tabular.

The agent may undergo one more retrieval step, leading to the selection of 4 new passages for each block. To maintain the quality of retrieved information and prevent excessive noise, we limit the maximum number of retrieval steps to three. This limitation is intended to avoid an accumulation of uncertainty, as additional passages introduce more information, making it more challenging for the retrieval process to distinguish between valuable information and noise.

Suppose the agent decides to generate the answer or has reached the maximum number of steps. In that case, the reader is presented with the question and all non-repeated passages (note that different blocks may contain the same passage). 
The reader utilizes the Reader Input to generate a predicted answer $a^{p}$, which is subsequently compared to the golden answer  \(a^g\). In cases where the reader is invoked before any passage has been retrieved, a preliminary Table Retrieval is run to ensure that a minimum of 10 documents are available for processing.

\subsection{Reward}

Since we lack prior knowledge of the correct sequence of actions for the agent, we cannot use supervised learning and must rely on the comparison between the expected (golden) answer and the agent's predicted answer to construct the reward function.

We started with the reward function \(R^3\), which trains the retriever indirectly, by using the final answer. However, due to the complexity of the problem, the -1 penalty for a generated answer with no intersection with the golden answer seemed excessive, especially considering that the Iterative-Retrieval + Cross-Block Reader, our inspiration, achieves an F1-score of only 20.9 on the test set. Therefore, we adjusted the penalty to -0.5. The following equation represents the reward at the end of the episode:

\begin{equation} \label{eq:reward}
    R(a^g, a^{rc} ) = 
    \begin{cases}
      2, \text{if } a^g = a^{rc}, \\
      F_1 (a^g, a^{rc}), \text{ else if } a^g \cap a^{rc} \neq \emptyset, \\
      -0.5, \text{else,}
    \end{cases}
\end{equation}
where the reward is assigned a value of 2 when the golden answer \(a^g\) precisely matches the predicted answer $a^{p}$. In cases of partial match, the reward is determined by the F1-score. Lastly, when no match occurs, the reward defaults to -0.5, as mentioned before.
We also introduced a minor penalty of -$0.02$ for each retrieval action taken. This penalty encourages the agent to minimize retrieving unnecessary texts and tables.

\subsection{Actions}
As previously mentioned, our approach involves three distinct actions: \textbf{ Retrieve Texts} ($A1$), where a retriever model searches for relevant textual information; \textbf{Retrieve Tables} ($A2$), where a retriever model searches for relevant tables; and \textbf{Generate Answer} ($A3$), where a reader model produces the answer using extracted textual passages and tables.

Next, we discuss our choices for the reader and retriever models. We opted not to retrain any models from the literature, limiting ourselves to those publicly available.

\textbf{Retriever:} the retriever module gathers information for the reader. We evaluated two types: BM25 \cite{Robertson2009} and Tri-encoder \cite{tri_encoder}. This approach allowed us to assess classical sparse retrieval and neural dense retrieval methods, akin to the Iterative-Retrieval + Cross-Block Reader architecture.

\textbf{Reader:} a critical component of any QA system, the reader generates answers. For OTT-QA, the publicly available Fusion-in-Encoder (FiE) model from the COS system \cite{cos} demonstrated superior performance. Notably, this model handles passages individually, making it versatile for handling multiple inputs. Following the authors' recommendations, we capped the number of input passages at 50, which yielded optimal results in our validation experiments.

\subsection{State/Observation}
The observation is the information that the agent uses for the decision-making process. In our case, this information comprises the question and the passages that have been retrieved up to that point. Rather than working directly with pure text, we first convert these elements into numerical data using embeddings.

We employ two distinct encoders to perform this transformation into embeddings. When the retriever is the Tri-encoder, we utilize its embeddings to represent the question, retrieved passages, and retrieved tables. However, in the case of BM25, we leverage the representations provided by a fine-tuned MPNET \cite{song2020mpnet} model trained for semantic search in QA\footnote{The MPNET model can be at \url{https://huggingface.co/sentence-transformers/multi-qa-mpnet-base-dot-v1}}. Each embedding is a vector with length of 512 for Tri-encoder (same as small-BERT) and 768 for the MPNET model.

This information is then translated into a sequence of 11 vectors, with one vector for the question and one for each of the 10 blocks. For each block, we calculate the average of the representation values for each passage, considering the question in isolation, generating the embedding sequence.

\subsection{Agent}
In DRL applications, the agent's primary function is to process observations and generate corresponding actions. Our specific observations consist of a sequence comprising 11 vectors. In our exploration, we experimented with four distinct neural network architectures: MultiLayer Perceptron (MLP), Long Short-Term Memory (LSTM), Gated Recurrent Unit (GRU), and Transformer.

For the MLP, we flattened the entire sequence into a single vector, which meant adding zeros in the first step to account for the lack of retrieved passages at that point. In contrast, the other networks process this information as a sequence.
 
\begin{figure}[t]
    \centering
    \includegraphics[width=1\textwidth]{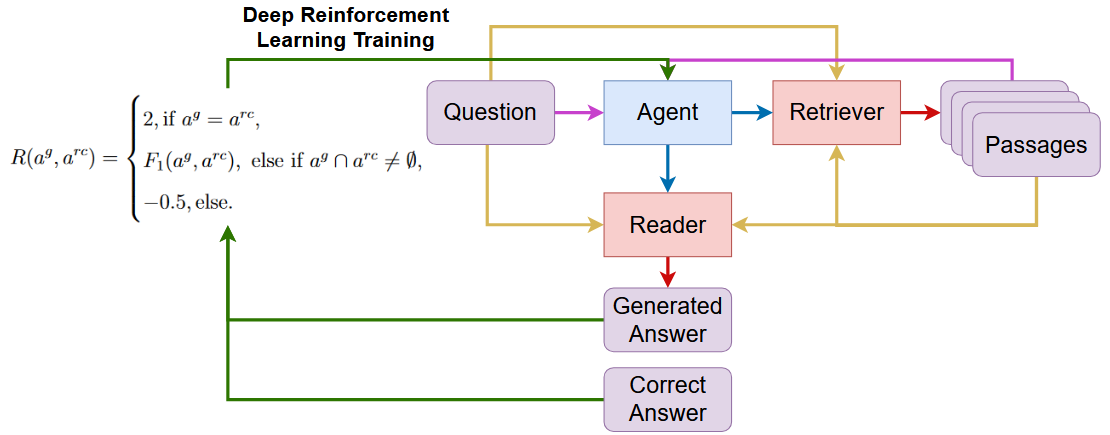}
    \caption[Training workflow.]{Training workflow of the proposed architecture.}
    \label{fig:training_workflow}
\end{figure}

To train the agent, we explored two widely-used algorithms: DQN \cite{DQN} and PPO \cite{PPO}. Our implementation is based on the Stable-Baselines3 Python library, selected for its robustness and reliability. 
The training workflow of the agent is depicted  in Figure \ref{fig:training_workflow}.

\section{Experiments}\label{chapter:experiments}

\subsection{Baselines}

Before training the DRL agent, we explored baseline scenarios using our architecture. These baselines involved fixed sets of actions chosen independently of the question and retrieved information. Each conceivable combination of actions was systematically evaluated across a validation dataset. For example, one baseline involved the agent consistently choosing to retrieve texts (Action $A1$) twice and then invoking the reader to generate the answer. In total, we explored 14 different configurations of actions, encompassing two retriever types—BM25 and Tri-encoder—resulting in 28 baselines. The primary purpose of these baseline tests was to establish a benchmark for comparison, determining if the trained DRL agent could outperform or at least match the performance of fixed action paths without contextual understanding.

\subsection{Training}

The training of the DRL agent involved sampling a random question from the OTT-QA training set for each episode. We evaluated the impact of training both retriever solutions and compared their performance. For the Tri-encoder, we trained for a total of one million timesteps, a reasonable number considering the dataset size of 41,469 training questions. Each question could take up to four timesteps (between retriever and reader), allowing for revisiting the same question multiple times. Due to the higher inference time of BM25 (up to 20 times more than Tri-encoder), we opted to train for only 100 thousand timesteps.
 
We implemented the DQN and PPO algorithms, employing various neural network architectures, including MLP, LSTM, GRU, and transformer. Subsequent subsections provide detailed explanations of these implementations.

\textbf{DQN:} For training DQN, we made a few adjustments to the dedefault hyperparameters of Stable-Baselines3,\footnote{\url{https://stable-baselines3.readthedocs.io}} primarily to account for the number of training steps. We set the buffer size to 500K (maximum number of transitions stored in the experience replay) and learning starts (number of steps taking random actions just to store some transitions in the buffer before training) to 50,000 for the system with Tri-encoder as the retriever. For BM25, we decreased the learning starts by a factor of 5, taking into account that the training has only 100,000 steps.

\textbf{PPO:} Similarly to DQN, we mostly relied on the default hyperparameters for training using the PPO implementation from Stable-Baselines3. We set the number of steps to run before training as 128, with a batch size of 32, and 60 as the number of epochs to optimize the surrogate loss.

As PPO is an Actor-Critic algorithm, it employs two different networks: the actor and the critic. Instead of using entirely separate networks, a common approach is to use a feature extractor to avoid redundant computations. This involves using shared layers between the networks for pre-processing the input, with these shared layers being trained jointly. 

\textbf{Networks:} Below are the configurations for the feature extractors we used, maintaining consistency for both training algorithms. For PPO, we employed two hidden layers with 64 and 32 neurons for both the actor and critic networks after the feature extractor:

\begin{itemize}
    \item \textbf{MLP:} for DQN, the MLP network is a simple network with three hidden layers having 512, 128, and 64 neurons. For PPO, the feature extractor is two-layered network with hidden sizes of 512 and 128.
        
    \item \textbf{LSTM:} we used 2 LSTM layers with a hidden size of 512 for the Tri-encoder and 768 for the BM25 architecture to match the size of the vectors of the embedding sequence $E$, with a dropout of 0.1. Additionally, there is one hidden layer with 128 nodes.
    
    \item \textbf{GRU:} similar to LSTM, but with the LSTM layers replaced by GRU layers.
    
    \item \textbf{Transformer:} for the PPO feature extractor, we opted for two encoder transformer layers with two attention heads, a feature dimension matching the input embedding size, and the same 128 hidden layer. For DQN, we used the same architecture with an extra 64 neurons last layer.

\end{itemize}

\section{Results and Discussion}\label{chapter:results}

\subsection{Baselines}
The complete results of our baselines are provided in Table \ref{tab:baselines}. For the BM25 retriever, it is evident that retrieving tables is less effective than retrieving texts. This imbalance results in the best sequence of actions always being the retrieval of texts and never tables, yielding an F1-score of 19.03. For the Tri-encoder, the results are more balanced. The optimal sequence of actions involves retrieving texts, followed by tables, and then retrieving texts again, resulting in a total F1-score of 8.24. It is notable, however, that the Tri-encoder performance when retrieving texts is considerably lower compared to BM25.

\begin{table}[ht!]  
\scriptsize
\centering
\begin{tblr}{
  width = \linewidth,
  colspec = {Q[160]Q[110]Q[110]Q[110]Q[110]Q[120]Q[180]},
  cells = {c},
  vline{2-7} = {-}{},
  hline{2-30} = {-}{},
}  
\textbf{Retriever} & \textbf{Action~1} & \textbf{Action~2} & \textbf{Action~3} & \textbf{Action~4} & \textbf{EM (\%)} & \textbf{F1-score (\%)} \\ \hline
BM25 & $A1$                & $A3$                & -                 & -                 & 12.92            & 16.80                  \\
BM25 & $A2$                & $A3$                & -                 & -                 & 2.17             & 4.20                   \\
BM25 & $A1$                & $A1$                & $A3$                & -                 & 14.60            & 18.79                  \\
BM25 & $A1$                & $A2$                & $A3$                & -                 & 13.15            & 16.88                  \\
BM25 & $A2$                & $A1$                & $A3$                & -                 & 5.01             & 7.97                   \\
BM25 & $A2$                & $A2$                & $A3$                & -                 & 2.12             & 4.07                   \\
\textbf{BM25} & \textbf{$A1$}       & \textbf{$A1$}       & \textbf{$A1$}       & \textbf{$A3$}       & \textbf{14.81}   & \textbf{19.03}         \\
BM25 & $A1$                & $A1$                & $A2$                & $A3$                & 14.23            & 18.21                  \\
BM25 & $A1$                & $A2$                & $A1$                & $A3$                & 13.32            & 17.36                  \\
BM25 & $A1$                & $A2$                & $A2$                & $A3$                & 7.59             & 10.52                  \\
BM25 & $A2$                & $A1$                & $A1$                & $A3$                & 5.15             & 8.08                   \\
BM25 & $A2$                & $A1$                & $A2$                & $A3$                & 3.61             & 6.17                   \\
BM25 & $A2$                & $A2$                & $A1$                & $A3$                & 3.21             & 5.45                   \\
BM25 & $A2$                & $A2$                & $A2$                & $A3$                & 2.08             & 4.15                   \\ 
Tri-encoder & $A1$                & $A3$                & -                 & -                 & 3.52             & 6.32                   \\
Tri-encoder & $A2$                & $A3$                & -                 & -                 & 2.94             & 4.41                   \\
Tri-encoder & $A1$                & $A1$                & $A3$                & -                 & 4.34             & 7.11                   \\
Tri-encoder & $A1$                & $A2$                & $A3$                & -                 & 4.83             & 7.54                   \\
Tri-encoder & $A2$                & $A1$                & $A3$                & -                 & 5.42             & 7.89                   \\
Tri-encoder & $A2$                & $A2$                & $A3$                & -                 & 2.66             & 4.40                   \\
Tri-encoder & $A1$                & $A1$                & $A1$                & $A3$                & 4.65             & 7.35                   \\
Tri-encoder & $A1$                & $A1$                & $A2$                & $A3$                & 4.79             & 7.56                   \\
\textbf{Tri-encoder} & \textbf{$A1$}       & \textbf{$A2$}       & \textbf{$A1$}       & \textbf{$A3$}       & \textbf{5.55}    & \textbf{8.24}          \\
Tri-encoder & $A1$                & $A2$                & $A2$                & $A3$                & 3.66             & 5.81                   \\
Tri-encoder & $A2$                & $A1$                & $A1$                & $A3$                & 5.51             & 8.06                   \\
Tri-encoder & $A2$                & $A1$                & $A2$                & $A3$                & 4.29             & 6.45                   \\
Tri-encoder & $A2$                & $A2$                & $A1$                & $A3$                & 4.07             & 6.24                   \\
Tri-encoder & $A2$                & $A2$                & $A2$                & $A3$                & 2.89             & 4.52                   
\end{tblr}
\caption{Baseline results for the OTT-QA validation dataset using our framework with BM25 and Tri-encoder as the retriever. In this experiment, we assumed that the agent always takes the same sequence of actions, regardless of the question. $A1$ corresponds to retrieving texts, $A2$ is retrieving tables, and $A3$ calls the reader to generate the answer.}
\label{tab:baselines}
\end{table}


\subsection{Results for Deep Reinforcement Learning Agent}

We present detailed results for the trained DRL agents:


\textbf{BM25} The performance of trained algorithms using the BM25 retriever is detailed in Table \ref{tab:trained}. Among the various training algorithms, the PPO agent equipped with a transformer neural network demonstrated remarkable results. Surpassing other configurations, this agent achieved the same level of performance as the best baseline, which involved consistently selecting the $A1$ action regardless of the question or context.


\textbf{Tri-encoder} Results by algorithms utilizing the Tri-encoder retriever are presented in Table \ref{tab:trained}. Notably, the best-performing agent emerged from the PPO algorithm, employing the MLP network, demonstrating a F1-score of 6.44\%.


\begin{table}[!ht]
\scriptsize
\centering
\begin{tblr}{
  width = \linewidth,
  colspec = {Q[210]Q[210]Q[190]Q[150]Q[237]},
  cells = {c},
  vline{2-5} = {-}{},
  hline{2-17} = {-}{},
} 
\textbf{Retriever} & \textbf{Algorithm} & \textbf{Network} & \textbf{EM (\%)} & \textbf{F1-score (\%)} \\ \hline
BM25 & DQN                         & MLP                  & 11.70            & 15.44                  \\
Bm25 & DQN                         & LSTM                 & 12.33            & 16.42                  \\
BM25 & DQN                         & GRU                  & 13.23            & 17.16                  \\
BM25 & DQN                         & Transformer          & 11.65            & 15.12                  \\
BM25 & PPO                         & MLP                  & 8.54             & 11.63                  \\
BM25 & PPO                         & LSTM                 & 6.78             & 9.85                   \\
BM25 & PPO                         & GRU                  & 11.25             & 14.90                  \\
\textbf{BM25} & \textbf{PPO}                & \textbf{Transformer} & \textbf{14.81}   & \textbf{19.03}  \\
Tri-encoder & DQN                         & MLP              & 3.61             & 6.01                   \\
Tri-encoder & DQN                         & LSTM             & 3.52             & 5.68                   \\
Tri-encoder & DQN                         & GRU              & 3.79             & \textbf6.13                   \\
Tri-encoder & DQN                         & Transformer      & 3.34             & 5.09                   \\
Tri-encoder & PPO                & MLP     & 3.38    & 5.47          \\
Tri-encoder & PPO                         & LSTM             & 3.75             & 5.53                   \\
Tri-encoder & PPO                         & GRU              & 2.94             & 4.93                   \\
Tri-encoder & PPO                         & Transformer      & 2.80             & 4.34                 
\end{tblr}
\caption{Results for   networks and training algorithms on the  OTT-QA dataset.}
\label{tab:trained}
\end{table}
\subsection{Discussion}

Baseline results show a significant advantage for text retrieval over table search with BM25. This may be due to BM25's difficulty in handling structured tables, where column presence and data types are crucial. Despite its popularity in OTT-QA solutions, the Tri-encoder achieved F1-scores below 10\%, potentially due to the lack of multi-hop training and reliance on text retrieval data from other datasets, as OTT-QA lacks this information.


The DRL agents showed performance comparable to or slightly below the best baseline, particularly for BM25, often favoring text retrieval, which aligns with the optimal baseline strategy. Table retrieval proved noisy, especially in iterative search, as tables often contain excessive data with relevant information typically in a single row. This issue may be mitigated by breaking tables into rows or smaller content sets, as suggested by other researchers \cite{core}.

The PPO algorithm generally showed slightly lower performance, with one exception: the PPO agent with a Transformer network for BM25, which achieved the best performance by converging to the trivial solution of only picking the best action on average (retrieving texts). The choice of neural network architecture had minimal impact, indicating the need for further experiments.

Our best-performing DRL agent achieved an F1-score of 19.03 using BM25, approaching the 20.7 score of the Iterative-Retrieval (sparse) + Cross-Block solution, which inspired our approach. However, our dense approach reached an F1-score of 8.24, compared to their 18.5, suggesting that the Tri-encoder setting requires further refinement.


Our solution has a few limitations. The exploratory nature of Reinforcement Learning (RL) and the stochastic behavior of neural networks benefit from averaged multiple test runs. However, due to time constraints—each training execution took over a week—we were not able to perform multiple runs. Additionally, we only used 100k training steps with the BM25 retriever. This number of steps is potentially insufficient given the dataset's 41,469 training questions, each potentially requiring up to three steps. Finally, on the Tri-encoder's performance, it is important to stress that this model was not trained for iterative retrieval, likely affecting its efficacy. 


\section{Conclusion and Future Work}\label{chapter:conclusion}

In this work, we introduced a novel system designed to tackle multi-hop questions across texts and tables. Our approach relies on a DRL decision-maker system that iteratively selects among three distinct modules: Text Retriever, Table Retriever and Reader. Our best system achieved an F1-score of 19.03, closely competing with the non-DRL Iterative-Retrieval (sparse) + Cross-Block system, which scored 20.7. 
We emphasize that our goal here is to explore sequential strategies; other (non-sequential) strategies
have been explored with success --- for instance, COS, which achieved an F1-score of 63.2 on the dev set. Despite the comparatively lower performance, we believe there is significant value in solutions that explore sequential decisions as they have potential for modular improvement by refining various system modules.  

Solutions that combine sequential decisions with non-sequential modules seem to be the most profitable avenue to explore in future work; here we have explored the performance boundary
of sequential decisions using reinforcement learning.
Future work should also include implementing the Joint-Reranking approach \cite{li2021dual}, which uses a BERT-based architecture to score and select top passages, enhancing the system's ability to clear excess passages and tables from memory. We are also exploring alternative methods for generating observations for the agent. Currently, we average the embeddings of each block, but we are considering encoding all distinct retrieved passages as a sequence. Additionally, integrating our DRL agent with the COS architecture, which excels in OTT-QA by employing diverse retrieval strategies, could enhance performance. The DRL agent would determine optimal retrieval paths based on the question and retrieved passages. 


\begin{credits}
\subsubsection{\ackname} 

This research has been carried out with  support by \textit{Ita\'{u} Unibanco S.A.} through the scholarship program  \textit{Programa de Bolsas Ita\'{u}} (PBI); it is also supported in part by the Coordenação de Aperfeiçoamento de Pessoal de Nível Superior (CAPES), Finance Code 001, Brazil.
The authors would like to thank the Center for Artificial Intelligence (C4AI-USP), with support by the São Paulo Research Foundation (FAPESP) under grant number 2019/07665-4 and by the IBM Corporation. Paulo Pirozelli is supported by the FAPESP grant 2019/26762-0. 

Any opinions, findings, and conclusions expressed in this manuscript are those of the authors and do not necessarily reflect the views, official policy or position of the Itaú-Unibanco, FAPESP, IBM, and CAPES.

\end{credits}
%
%
%
\bibliographystyle{splncs04}
\bibliography{References}

\end{document}